\documentclass[10pt,twocolumn,letterpaper]{article}

\usepackage[pagenumbers]{iccv} 
\usepackage{graphicx}
\usepackage{amsmath}
\usepackage{booktabs}
\usepackage{amssymb}
\usepackage{arydshln}
\usepackage{booktabs} 
\usepackage[accsupp]{axessibility}

\newcommand{\our}{{PhysSplat}\xspace}

\usepackage{epsfig}
\usepackage{xcolor}
\usepackage{graphicx}
\usepackage{float}
\usepackage[utf8]{inputenc}

\usepackage{caption}
\usepackage{subcaption}
\usepackage{array}
\usepackage{algorithm}
\usepackage{algpseudocode}
\usepackage{multirow}
\usepackage{etoolbox}
\usepackage{anyfontsize}
\usepackage{booktabs}
\usepackage{colortbl}
\usepackage{bm}
\usepackage{pifont}
\usepackage{adjustbox}
\usepackage[accsupp]{axessibility}

\newcommand{\cmark}{\textcolor{green}{\ding{51}}} 
\newcommand{\ccmark}{\cmark\kern-0.4em\cmark} 
\newcommand{\xmark}{\textcolor{red}{\ding{55}}} 

\newcolumntype{R}[2]{%
    >{\adjustbox{angle=#1,lap=\width-(#2)}\bgroup}%
    l%
    <{\egroup}%
}
\newcommand*\rot{\multicolumn{1}{R{35}{1em}}}

\definecolor{iccvblue}{rgb}{0.21,0.49,0.74}
\usepackage[pagebackref,breaklinks,colorlinks,allcolors=iccvblue]{hyperref}

\usepackage[capitalize]{cleveref}
\crefname{section}{Sec.}{Secs.}
\Crefname{section}{Section}{Sections}
\Crefname{table}{Table}{Tables}
\crefname{table}{Tab.}{Tabs.}

\def\paperID{4207} 
\def\confName{ICCV}
\def\confYear{2025}

\title{PhysSplat \raisebox{-0.5\baselineskip}{\includegraphics[scale=0.038]{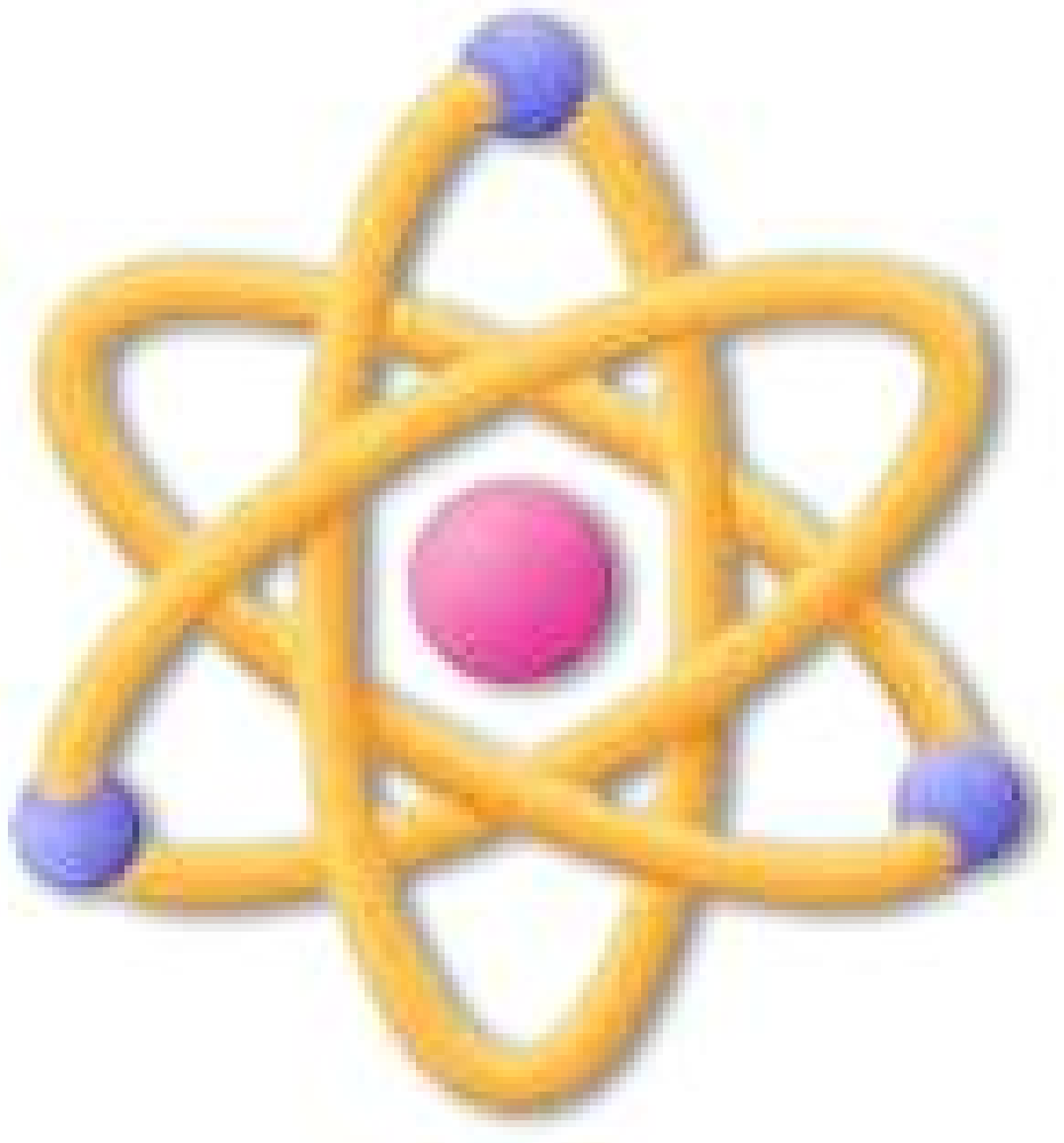}}: Efficient Physics Simulation for 3D Scenes \\ via MLLM-Guided Gaussian Splatting}

\author{
Haoyu Zhao\textsuperscript{* 1} \quad 
Hao Wang\textsuperscript{* 2} \quad
Xingyue Zhao\textsuperscript{* 4} \quad
Hao Fei\textsuperscript{5} \quad \\
Hongqiu Wang\textsuperscript{6} \quad
Chengjiang Long\textsuperscript{\textdagger 3} \quad
Hua Zou\textsuperscript{\textdagger 1}\\
\textsuperscript{1}School of Computer Science, Wuhan University \quad \\
\textsuperscript{2}Wuhan National Laboratory for Optoelectronics, Huazhong University of Science and Technology \quad \\
\textsuperscript{3}Meta Reality Lab \quad
\textsuperscript{4}Xi’an Jiao Tong University \quad
\textsuperscript{5}National University of Singapore \\
\textsuperscript{6}The Department of Systems Hub, Hong Kong University of Science and Technology (Guangzhou)\\
}

\begin{document}

\twocolumn[{%
    \renewcommand\twocolumn[1][]{#1}%
    \setlength{\tabcolsep}{0.0mm} 
    \newcommand{\sz}{0.125}  
    \maketitle
    \begin{center}
        \newcommand{\teaserwidth}{\textwidth}
        \includegraphics[width=\linewidth]{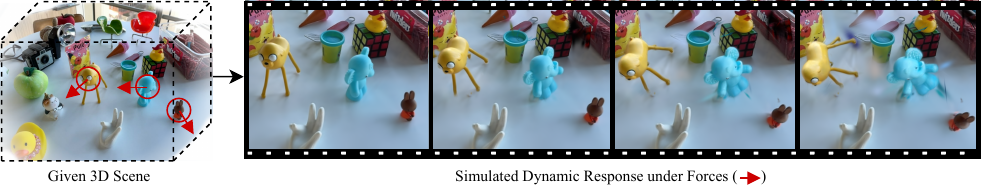}
    \vspace{-23pt}
    \captionof{figure}{We develop an efficient method for simulating the dynamic movements of 3D objects with customizable behaviors, and synthesizing interactive 3D dynamics under arbitrary forces (red arrows).
    Compared to recent methods~\cite{zhang2024physdreamer,liu2024physics3d,huang2024dreamphysics}, our approach produces more realistic 3D dynamics with much faster inference times.} 
    \label{fig:teaser}
    \end{center}%
}]

\maketitle
{
\renewcommand{\thefootnote}{\fnsymbol{footnote}}
\footnotetext{* Equal contributions.}
\footnotetext{\textdagger Corresponding Author.}
}

\begin{abstract}
Recent advancements in 3D generation models have opened new possibilities for simulating dynamic 3D object movements and customizing behaviors, yet creating this content remains challenging.
Current methods often require manual assignment of precise physical properties for simulations or rely on video generation models to predict them, which is computationally intensive.
In this paper, we rethink the usage of multi-modal large language model (MLLM) in physics-based simulation, and present \textbf{\our}, a physics-based approach that efficiently endows static 3D objects with interactive dynamics.
We begin with detailed scene reconstruction and object-level 3D open-vocabulary segmentation, progressing to multi-view image in-painting. 
Inspired by human visual reasoning, we propose MLLM-based Physical Property Perception (MLLM-P3) to predict the mean physical properties of objects in a zero-shot manner. 
The Material Property Distribution Prediction model (MPDP) then estimates physical property distributions via geometry-conditioned probabilistic sampling of MLLM-P3 outputs, reformulating the problem as probability distribution estimation to reduce computational costs. 
Finally, we simulate objects in 3D scenes with particles sampled via the Physical-Geometric Adaptive Sampling (PGAS) strategy, efficiently capturing complex deformations and significantly reducing computational costs.
Extensive experiments and user studies demonstrate that our \our achieves more realistic motion than state-of-the-art methods \textbf{within 2 minutes} on a single GPU.
Here is our \href{https://sim-gs.github.io/}{project page}.
\end{abstract}

\section{Introduction}
\label{sec:intro}

With the development in 3D representation, Neural Radiance Fields (NeRF)~\cite{mildenhall2021nerf} and 3D Gaussian Splatting (3DGS)~\cite{kerbl20233d} offer new perspectives for 3D reconstruction and 3D representation~\cite{wang2024prolificdreamer,tang2023dreamgaussian,zhao2024hfgs}. However, these approaches are unable to simulate interactions with 3D assets in simulation environments~\cite{savva2019habitat,xia2018gibson}, which is s critical for generating realistic object responses to novel interactions, such as external forces or agent manipulations in many applications, \text{e.g.}, virtual reality~\cite{jiang2024vr}, embodied intelligence~\cite{lu2024manigaussian}.

\begin{table}[t]
    \begin{center}
    \small
    \setlength{\tabcolsep}{8.5pt}  
    \setlength{\belowcaptionskip}{-15pt}
    \begin{tabular}{ccccc|l}
    \rot{auto param computation} 
    & \rot{fast inference time}  
    & \rot{physics-based deformation} 
    & \rot{static 3D object input}  
    & \rot{scene-wide simulation} &  \\
   \toprule
    \rowcolor{gray!10}\xmark & \cmark & \cmark & \cmark & \xmark & PhysGaussian~\cite{xie2024physgaussian} \\
    
    \xmark & \xmark & \xmark & \cmark & \xmark & DreamGaussian4D~\cite{ren2023dreamgaussian4d} \\

    \rowcolor{gray!10}\xmark & \xmark & \xmark & \cmark & \xmark & Animate124~\cite{zhao2023animate124} \\

    \xmark & \cmark & \cmark & \xmark & \xmark & PAC-NeRF~\cite{li2023pac} \\

    \rowcolor{gray!10}\xmark & \cmark & \cmark & \xmark & \xmark & PIE-NeRF~\cite{feng2024pie} \\

    \xmark & \cmark & \cmark & \xmark & \xmark & Spring-Gaus~\cite{zhong2024reconstruction} \\

    \rowcolor{gray!10}\cmark & \xmark & \cmark & \cmark & \xmark & DreamPhysics~\cite{huang2024dreamphysics} \\

    \cmark & \xmark & \cmark & \cmark & \xmark & PhysDreamer~\cite{zhang2024physdreamer} \\

    \rowcolor{gray!10}\cmark & \xmark & \cmark & \cmark & \xmark & Physics3D~\cite{liu2024physics3d} \\

    \xmark & \xmark & \cmark & \cmark & \cmark & Feature Splatting~\cite{qiu2024feature} \\
    \hline
    \rowcolor[HTML]{D7F6FF}\cmark & \cmark & \cmark & \cmark & \cmark & \textbf{\our}\\

    \end{tabular}
    \end{center}
      \vspace{-14pt}
    \caption{\textbf{Comparison to Concurrent Works.} Our method is the only one that can simulate the entire scene at a much faster speed.}
  \vspace{-10pt}
    \label{tab:comparison_concurrent}
\end{table}

Some recent approaches aim to bridge the gap between rendering and simulation integrating physics-based priors into 3D object representations using physical simulators~\cite{chen2022virtual,qiu2024feature,feng2024pie}. For instance, PAC-NeRF~\cite{li2023pac} estimates the geometry and physical parameters of objects from multi-view videos and then integrates physical models with NeRF-based representations.
Similarly, PhysGaussian~\cite{xie2024physgaussian} first injects physical parameters into 3DGS objects and then predicts motion using a physics-based simulator. However, both methods are limited in handling real objects, as they require predefined material models or multi-view videos to predict physical properties.

To automatically set parameters, some approaches~\cite{liu2024physics3d,huang2024dreamphysics,zhang2024physdreamer} leverage video generation models~\cite{blattmann2023stable} that are trained on real-world video data to estimate the parameters of physical materials.erials. For example, PhysDreamer~\cite{zhang2024physdreamer} employs a stable video diffusion model to learn Young's modulus of objects.
However, learning material physical properties from video diffusion priors is computationally expensive and time-consuming in practice. Moreover, video diffusion models have limited controllability and often fail to obey physical laws~\cite{ren2023dreamgaussian4d,zhao2023animate124}. Additionally, these models are also generally restricted to non-rigid objects, making them unsuitable for deriving the physical properties of large rigid objects (such as cup, bowl, and chairs). 
However, humans are exceptionally adept at predicting the physical properties of objects based on visual information~\cite{fleming2014visual,fleming2013perceptual}. We therefore ask this question: \textit{how can we develop models for perceiving physics from just visual data}? 

To this end, we rethink physics-based simulation and the usage of multi-modal large language model (MLLM), such as GPT-4V~\cite{yang2023dawn}. In this paper, we propose \textbf{\our}, a physics-based method that efficiently transforms static 3D objects into interactive ones capable of responding to new interactions, as shown in Fig.~\ref{fig:teaser}. We first segment objects with priors from foundation models~\cite{liu2023grounding,kirillov2023segment,zhang2024recognize}.
Inspired by how humans predict physical properties of objects through visual data, \our leverages MLLM-based Physical Property Perception (MLLM-P3) to predict the mean values of physical properties. Unlike previous methods~\cite{zhang2024physdreamer,liu2024physics3d,huang2024dreamphysics} iteratively refining physical properties through video analysis, we reformulate this problem to a probability distribution estimation task by predicting the full range of these properties based on the mean value and the object's geometry, reducing computational demands.
Finally, \our simulates object interactions in a 3D scene with driving particles sampled by the Physical-Geometric Adaptive Sampling (PGAS) strategy.
Extensive experiments and user studies demonstrate that \textit{\our achieves more accurate physical property prediction and synthesizes more realistic motion with much faster inference time.} 
We provide an overview of the comparison to previous works in Tab.~\ref{tab:comparison_concurrent}. In summary, \our makes the following contributions:
\begin{itemize}
    \item \our is the first to use MLLM for zero-shot physical property estimation of objects in 3D scenes.
    \item We reformulate physical property estimation as a probability distribution task, enabling adaptable physical simulations with PGAS in open-world scenes.
    \item Experiments show \our effectively predict physical properties and creates realistic 3D dynamics.
\end{itemize}

\section{Related Work}
\label{sec:work}

\subsection{Dynamic 3D Animation}
The demand for dynamic 3D animation creation has grown significantly across various applications, including video games, virtual reality, and robotic simulation~\cite{zhao2024sg,zhao2024chase,zhao2025smap}. With the success of video generative models~\cite{wang2025serp}, some methods~\cite{zhao2023animate124} have attempted to leverage video diffusion models to guide the prediction of 3D deformations. For instance, DreamGaussian4D~\cite{ren2023dreamgaussian4d} uses pre-generated videos to supervise the deformation of static scenes. However, the deformations produced by these methods may not always be accurate or physically plausible.

Recent works~\cite{modi2024simplicits,zhong2024reconstruction} introduce physics simulation to the 3D deformation and enable synthesizing motions under any physical interactions. Virtual Elastic Objects~\cite{chen2022virtual} jointly reconstructs the geometry, appearances, and physical parameters of elastic objects with multi-view data. Spring-Gaus~\cite{zhong2024reconstruction} integrate a 3D Spring-Mass model into 3D Gaussian kernels, and then simulate elastic objects from videos of the object from multiple viewpoints.
PAC-NeRF~\cite{li2023pac} and PhysGaussian~\cite{xie2024physgaussian} integrate physics-based simulations with NeRF~\cite{mildenhall2021nerf} and 3DGS~\cite{kerbl20233d}, respectively, to model the deformation of elastic objects. However, these methods either require manual setup of physical properties for 3D objects before simulation or depend on multi-view videos to predict physical properties.

To avoid manually setting parameters, some works estimate physical material parameters with video generation model~\cite{blattmann2023stable} to estimate physical material parameters~\cite{huang2024dreamphysics}. PhysDreamer~\cite{zhang2024physdreamer} and DreamPhysics~\cite{huang2024dreamphysics} leverage video generation models to estimate physical material parameter (\text{e.g.}, Young's modulus), while Physics3D~\cite{liu2024physics3d} further optimizes a wider range of physical parameters for 3D objects. However, these methods are computationally expensive, as learning material properties through video diffusion priors is time-consuming. Moreover, the controllability of generated videos is limited, often deviating from physical laws~\cite{ren2023dreamgaussian4d,zhao2023animate124}, which we further demonstrate in the experimental Section~\ref{sec:exp}.
Additionally, these models are typically restricted to non-rigid objects, making them unsuitable for determining the physical properties of large rigid objects, such as tables, chairs, and sofas. Inspired by how humans perceive physical properties of the objects ~\cite{fleming2014visual,fleming2013perceptual}, we propose leveraging multi-modal large language models (MLLMs) to efficiently predict the mean values of physical properties for objects in a 3D scene, enabling faster inference times. We then use the proposed MPDP to predict the full distribution of these properties.

\subsection{Visual Physics Perception}
Physics perception is a long-standing challenging problem~\cite{wu2015galileo}. Previous studies demonstrate that deep learning models can potentially exhibit physical perception abilities similar to humans~\cite{bell2015material,hu2011toward,wang2024video}. Most prior research focuses on dynamically addressing object properties, either by observing the target’s behavior~\cite{li2023pac} or by interacting with it in a 3D physical engine~\cite{pinto2016curious,yao2023estimating}. Other works also explore the estimation of material properties directly from static images~\cite{bell2015material,sharan2013recognizing}. However, these works mostly focus on specific material properties, such as mass or tenderness, often relying on task-specific data. In contrast, we propose leveraging MLLM, such as GPT-4V~\cite{yang2023dawn}, to generate a wide range of physical properties such as mass, Young’s modulus, and Poisson's ratio in a zero-shot manner.

\section{Preliminaries}
\label{sec:pre}

\subsection{Material Point Method}


The Material Point Method (MPM)\cite{hu2018moving} is a popular simulation framework for multi-physics phenomena due to its capability to handle topology changes and frictional interactions. Unlike mesh-based methods, MPM represents the continuum using particles in a grid-based space, making it well-suited for point-based 3D Gaussian representation. Following PhysGaussian\cite{xie2024physgaussian}, we define each Gaussian kernel’s time-dependent state as:
\begin{equation}
x_i(t) = \Delta (x_i, t), \ \Sigma_i(t) = F_i(t) \Sigma_i F_i(t)^T,
\end{equation}
where $\Delta(\cdot,t)$ and $F_i(t)$ denote coordinate deformation and deformation gradient at timestep $t$. The viewpoint must also adjust with the continuum rotation $\Omega_i(t)$ to match the view direction of the spherical harmonic coefficient $C_i$.


\subsection{3D Gaussian Splatting (3DGS)}

3D Gaussian Splatting (3DGS) represents scenes as point clouds, with each point modeled as a 3D Gaussian defined by a center point $\mathcal{X}$ (mean) and a covariance matrix $\Sigma$. Each Gaussian at $\mathcal{X}$ is given by $G(\mathcal{X})=e^{-\frac{1}{2}\mathcal{X}^T\Sigma^{-1}\mathcal{X}}$. $\Sigma$ is decomposed into a scaling matrix $\mathcal{S}$ and rotation matrix $\mathcal{R}$, such that $\Sigma = \mathcal{R}\mathcal{S}\mathcal{S}^T\mathcal{R}^T$, with $\mathcal{S}$ and $\mathcal{R}$ stored as vectors $s \in \mathbb{R}^{N \times 3}$ and $r \in \mathbb{R}^{N \times 4}$, respectively. Differential splatting~\cite{yifan2019differentiable} applies a viewing transform $W$ and Jacobian $J$ to compute the transformed covariance $\Sigma^{\prime} = JW\Sigma W^TJ^T$, enabling novel view rendering. Each pixel color $\mathcal{C}$ is obtained by blending $N$ overlapping points:
\begin{equation}
\label{equa:gaussian_render}
\mathcal{C} = \sum_{i\in N}c_i \alpha_i \prod_{j=1}^{i-1} (1-\alpha_j),
\end{equation}
where $c_i$ and $\alpha_i$ denote color and opacity, derived from the Gaussian with covariance $\Sigma$ and optimized parameters.


\begin{figure*}[ht]
  \centering
    \includegraphics[width=\linewidth]{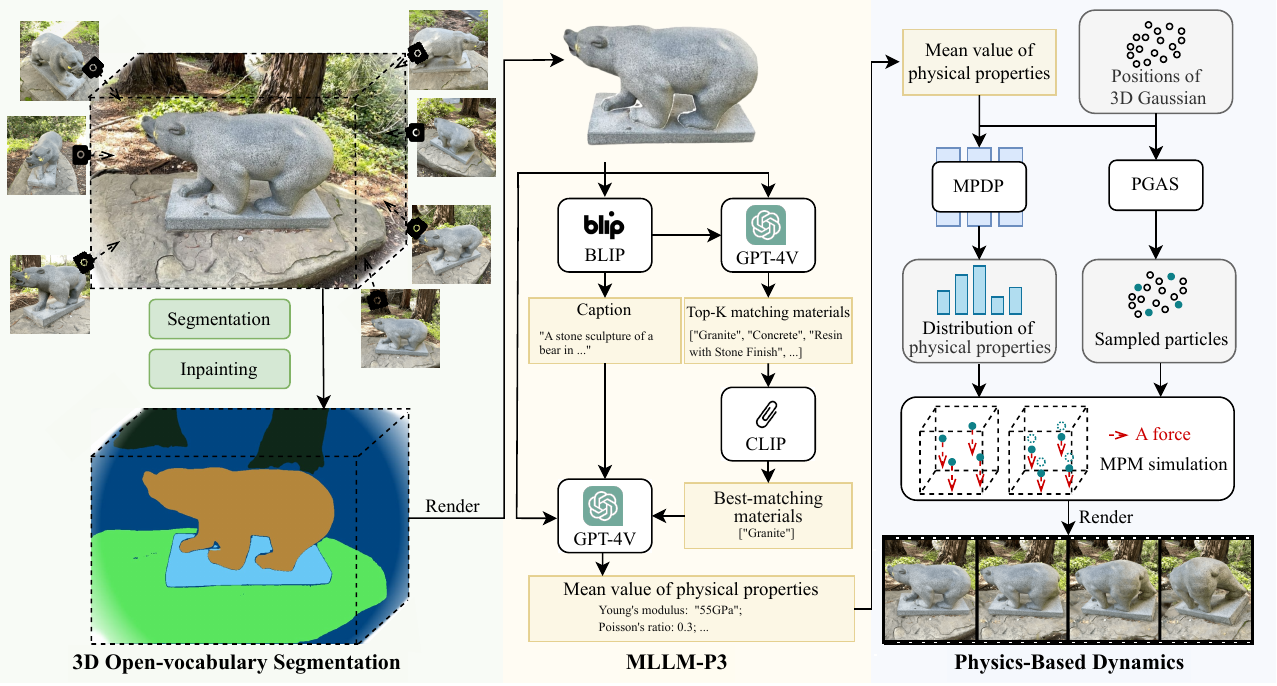}\\
  \vspace{-12pt}
  \caption{\textbf{Overview of \our \raisebox{-0.5\baselineskip}{\includegraphics[scale=0.025]{Fig/logo.pdf}}}.Given a pre-trained 3D scene and its corresponding 2D images, we first perform object-level segmentation of the 3D scene with the prior from a set of foundation meodels~\cite{liu2023grounding,kirillov2023segment,zhang2024recognize}. We obtain the mean physical properties of the object from the proposed MLLM-P3, and based on this and the object’s geometry, we then derive the full distribution using the MPDP model. Finally, we animate the 3D objects using a physics-based simulator with driving particles sampled via the Physical-Geometric Adaptive Sampling (PGAS) strategy.}
  \label{fig:pipeline}
\end{figure*}

\section{Our Methodology}
\label{sec:met}
Predicting various physical properties of 3D objects from static scene is an extremely challenging task due to limited supervisions. Instead of capturing physical data from generation models or multi-view videos~\cite{zhang2024physdreamer,liu2024physics3d,huang2024dreamphysics,li2023pac,zhong2024reconstruction,feng2024pie}, we reformulate this task from a new perspective, decomposing it into a set of sub-tasks. 
Specifically, as shown in Fig.~\ref{fig:pipeline}, we first segment the images with a set of foundation models~\cite{liu2023grounding,kirillov2023segment,zhang2024recognize} and lift these 2D segmented masks to segment 3D object in the scene via radiance fields rendering (Section~\ref{sec:segment}). We propose MLLM-based Physical Property Perception (MLLM-P3) to predict the mean values of these properties (Section~\ref{sec:gpt}). We then use the Material Property Distribution Prediction (MPDP) model to estimate the full distribution, simulating object dynamics with driving particles sampled using the Physical-Geometric Adaptive Sampling (PGAS) strategy (Section~\ref{sec:dynamic}).


\subsection{3D Open-vocabulary Segmentation}
\label{sec:segment}
For each scene, we first train a 3DGS model on given images and camera poses. Inspired by prior work~\cite{ren2024grounded}, we integrate 2D open-vocabulary models like Grounding DINO~\cite{liu2023grounding} for detection, SAM~\cite{kirillov2023segment} for segmentation, and RAM~\cite{zhang2024recognize} for tagging. These models automatically segment objects in images without textual input. Specifically, we use RAM to tag the image, Grounding DINO to create bounding boxes based on tags, and SAM to refine these boxes into precise masks. This approach enables full automatic image labeling using expert models.

After 2D open-vocabulary segmentation, each segmented image contains semantic features for each object. We project these 2D masks into 3D space using radiance field rendering. Inspired by recent work~\cite{zhao2024sg,ye2023gaussian}, each Gaussian retains its original attributes, with a learnable semantic attribute added for encoding object semantics. Using a zero-shot tracker~\cite{cheng2023tracking}, we assign unique IDs to masks across views, helping distinguish categories within the 3D scene through differentiable rendering (see Fig.\ref{fig:pipeline}). Extracting objects from 3DGS introduces holes, which we inpaint using LaMa\cite{suvorov2022resolution} to guide 3D Gaussian inpainting, keeping Gaussians outside holes fixed.





\subsection{MLLM-based Physical Property Perception}
\label{sec:gpt}
The variety of materials is vast and often indistinguishable by appearance alone, yet humans can infer material composition through high-level reasoning and visual cues. Recent research~\cite{driess2023palm} shows that multi-modal large language models (MLLM) excel in reasoning and decision-making for complex tasks. Inspired by human reasoning, we propose MLLM-based Physical Property Perception (MLLM-P3), which uses MLLM for open-vocabulary semantic reasoning about materials and their physical properties.

The segmented 3D scene in Section~\ref{sec:segment} is tightly related to the physical properties of the 3D objects in it. We first select a canonical view and render an object in 3D scene based on the 3D Gaussian's semantic attribute introduce Section~\ref{sec:segment}. Then we use a VQA model, such as BLIP~\cite{li2022blip} to produce a text description of the image. This description, along with the image, are then passed to a Multi-modal Large Language Model (MLLM) such as GPT-4V~\cite{yang2023dawn}, prompting it to return a dictionary containing K candidate materials and information on whether the object is rigid (related to the sampling method in Section~\ref{sec:dynamic}). To address potential hallucination, we compute the CLIP~\cite{radford2021learning} similarity score between the image and the materials in the dictionary to select the most matching material name. Finally, the selected material name, image, and text description provide a structured input to the MLLM, grounding its outputs in a reliable context. This enables the model to return a list of physical properties for the object, $M = {\rho, E, \nu}$, where $\rho$ represents the density, $E$ is Young’s modulus, and $\nu$ is Poisson’s ratio. These material properties are essential for understanding an object’s motion under forces.


\begin{figure}[t]
  \centering
    \includegraphics[width=0.95\linewidth]{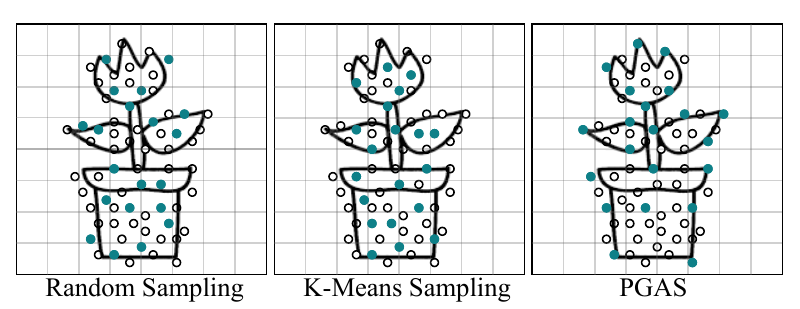}\\
  \vspace{-7pt}
  \caption{\textbf{Sampling}. We design a novel Physical-Geometric Adaptive Sampling (PGAS) strategy that captures the boundary of the object well. We employ PGAS to sample some “driving particles” (in green) and simulate only these particles. For rendering, each particle’s position and rotation are derived by fitting a local rigid body transformation based on neighboring driving particles.}
  \label{fig:sampele}
\end{figure}


\subsection{Physics-Based Dynamics}
\label{sec:dynamic}

\medskip
\noindent
\textbf{Material Property Distribution Prediction.}
Even for object composed of a single material, local physical properties exhibit inherent variations across different regions of the object~\cite{castaneda1995effect}. Additionally, the physical properties estimated by multi-modal large language model (MLLM) may not capture the 3D structure of the object. To address these challenges, we propose material property distribution prediction (MPDP), and reformulate the problem from a regression task to a probability distribution estimation task.

We train a network $\mathcal{D_\theta}$ using part of the synthesized dataset, with the object’s point cloud and predicted mean values (Section~\ref{sec:gpt}) as input. The network is supervised by the physical properties of particles predicted by Physics3D~\cite{liu2024physics3d}. Instead of using the raw Physics3D~\cite{liu2024physics3d} results, we normalize them by their mean value, allowing our MPDP to learn the geometry-based distribution of physical properties. With the more reliable mean values from MLLM-P3, our final estimates of physical properties are more accurate. Moreover, our approach is only about 2\% as time-efficient as Physics3D~\cite{liu2024physics3d}. 

The network predicts the geometry-aware probability distribution $\mathcal{P}$ of physical properties across particles, $\mathcal{P} = \mathcal{D_\theta}(\mathcal{X})$.
where $\mathcal{X}$ is the position of 3D Gaussians of the object. We then scale the distribution $\mathcal{P}$ by a global mean value predicted by the MLLM in Section~\ref{sec:gpt} through element-wise multiplication, yielding the final physical property values for each point in the material field. This approach efficiently estimates per-point physical attributes, such as Young’s modulus and Poisson’s ratio, across the entire point cloud while avoiding the computational overhead of per-particle calculations.


\medskip
\noindent
\textbf{Simulation with Physical-Geometric Adaptive Sampling.}
Rendering high-fidelity 3D scene often needs  millions of 3D Gaussians, which is significant computational demands for simulation. To reduce this burden, we implement a sub-sampling approach. Specifically, we design a Physical-Geometric Adaptive Sampling (PGAS) strategy. The original Poisson disk sampling requires that the distance between any two particles be larger than a threshold $r$. Starting from an initial point, PDS then tries to fill a banded ring between $r$ and 2$r$ with new samples. 

Our observation is that softer objects and those with complex shapes require more driving particles to accurately simulate their dynamics. To this end, we adaptively adjust the sample radius $r$ based on the object's Young’s modulus $E$ predicted in Section~\ref{sec:gpt} and curvature $K$. The curvature $K$ is defined as:
\begin{align}
C = \frac{1}{n} \sum_{j=1}^{n} & (\mathcal{X}_j - \bar{\mathcal{X}})(\mathcal{X}_j - \bar{\mathcal{X}})^T, \\
K = & \frac{\lambda_3}{\lambda_1 + \lambda_2 + \lambda_3},
\end{align}
where $\mathcal{X}_j$ is the position of the $j$-th 3D Gaussian of the object, $\bar{\mathcal{X}}$ is the mean position of all 3D Gaussians, and $\lambda_1, \lambda_2, \lambda_3$ are the eigenvalues of the covariance matrix $C$. Then, the sample radius $r$ is adjusted as:
\begin{align}
\hat{K} = \min&(V_{max}, \max(V_{min}, K)),\\
 \hat{r} = & \min(r,k \sqrt{\frac{E}{\hat{K}}}r), 
\label{eq:pgas}
\end{align}where we set $V_{max}=10$, $V_{min}=1$, and $k=\sqrt{10}$ in our paper. Our sampling ensures that the distance between a particle and its nearest neighbor is at least $\hat{r}$. By using smaller radii for softer materials and high-curvature areas, PGAS captures fine details more accurately, enhancing model resolution in deformation simulations and complex surface reconstruction, as shown in Fig.~\ref{fig:sampele}.


\medskip
\noindent
\textbf{MPM-Driven Physics-Based Dynamics.}
To model physical properties, we employ MLS-MPM~\cite{hu2018moving} as our simulator. In MPM, a continuum is represented by particles distributed in a grid-based space, offering a distinct advantage over mesh-based methods. MPM can be seamlessly applied to 3D Gaussian Splatting (3DGS) in point-based representations. Building on PhysGaussian~\cite{xie2024physgaussian}, we define a time-dependent state for each Gaussian kernel as follows:
\begin{equation}
x_i(t) = \Delta(x_i, t), \quad \Sigma_i(t) = F_i(t) \Sigma_i F_i(t)^T,
\end{equation}
where $\Delta(\cdot, t)$ and $F_i(t)$ represent the coordinate deformation and deformation gradient at time $t$. Additionally, to account for continuum rotation $\Omega_i(t)$, the rendering viewpoint is adjusted to align with the view direction of the spherical harmonic coefficient $C_i$.


\begin{figure*}[ht]
  \centering
    \includegraphics[width=\linewidth]{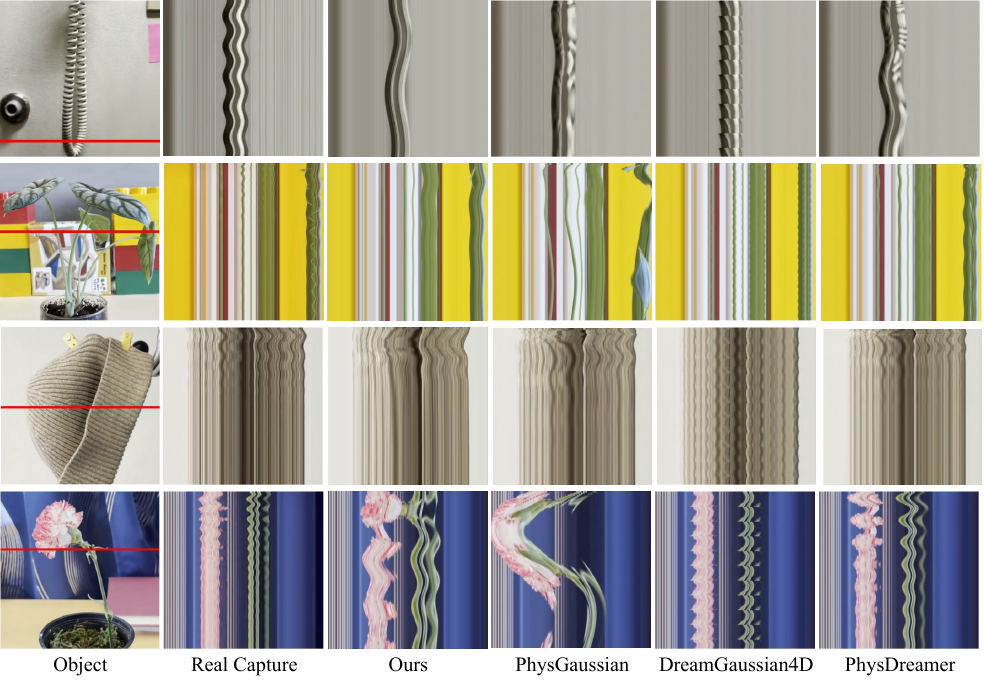}\\
  \vspace{-9pt}
  \caption{\textbf{Qualitative Comparison on PhysDreamer~\cite{zhang2024physdreamer}}. We compare our results with real captured videos, and some recent SOTA methods~\cite{xie2024physgaussian,zhang2024physdreamer,liu2024physics3d,ren2023dreamgaussian4d}. Our \our produces more realistic damping, closely matching real-world capture.}
  \vspace{-7pt}
  \label{fig:results_dreamer}
\end{figure*}


\begin{figure*}[ht]
  \centering
    \includegraphics[width=0.96\linewidth]{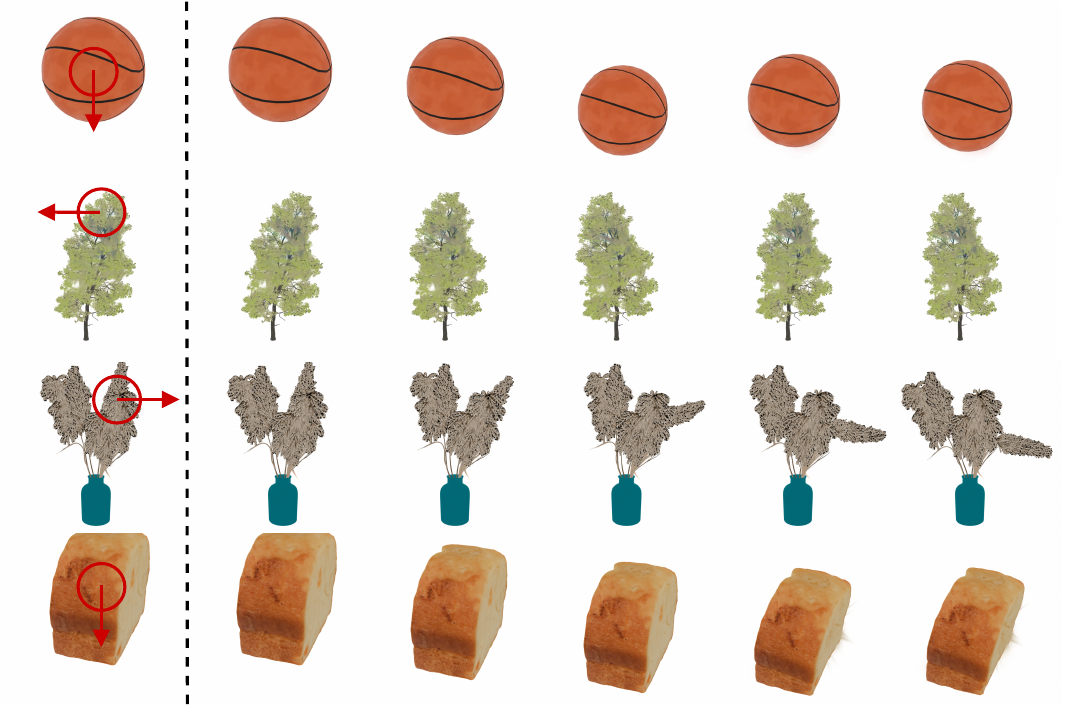}\\
  \vspace{-9pt}
  \caption{\textbf{Visual results on synthesized dataset~\cite{liu2024physics3d}} with an external force (red arrows). \our is able to generate realistic scene movement while maintaining good motion consistency.}
  \vspace{-7pt}
  \label{fig:results_3d}
\end{figure*}

\begin{figure}[t]
  \centering
    \includegraphics[width=1\linewidth]{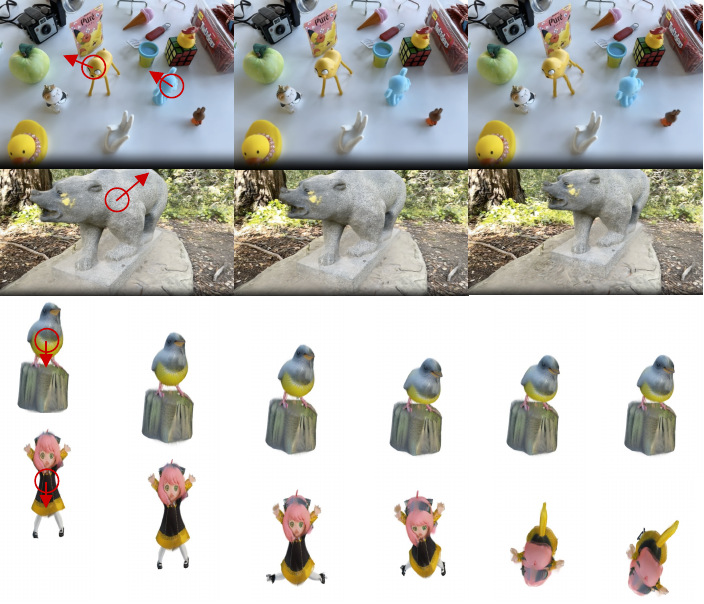}\\
  \vspace{-9pt}
  \caption{\textbf{Visual results} on Open-world dataset and 3D assets generated by LGM~\cite{tang2025lgm}.}
  \label{fig:rebuttal}
\end{figure}


\section{Experiments}
\subsection{Implementation Details}
We initiate the process by reconstructing 3D Gaussians from multi-view images and execute internal particle filling operations to refine the representation further. Each Gaussian kernel is then associated with a set of physical properties targeted for optimization following~\cite{zhang2024physdreamer,xie2024physgaussian}. We then discretize the foreground region into a grid structure, typically sized at $64^3$. For the MPM simulation, we use 768 sub-steps per interval between video frames, resulting in a sub-step duration of $4.34 \times 10^{-5}$ seconds to ensure precision and accuracy in simulation dynamics. All experiments are conducted on a single NVIDIA 3090 GPU.

\subsection{Datasets}
\medskip
\noindent
\textbf{PhysDreamer~\cite{zhang2024physdreamer}.} We also conduct experiments on the physical simulation of single objects on four real-world static scenes from PhysDreamer~\cite{zhang2024physdreamer} for fair comparison. Each scene includes an object and a background. 

\medskip
\noindent
\textbf{Synthesized dataset~\cite{liu2024physics3d}.} Following~\cite{liu2024physics3d}, we utilize BlenderNeRF~\cite{Raafat_BlenderNeRF_2023} to synthesize several scenes. Five cases are used to train the proposed MPDP model (as introduced in Section~\ref{sec:dynamic}), while the remaining four cases are reserved for subsequent comparisons.



\begin{table}[t]
 \centering
 \setlength{\tabcolsep}{12pt}  
 \begin{tabular}{lccc}
 \toprule
 Method & RS & AS & Time \\ 
 \hline
 \multicolumn{4}{c}{\textbf{PhysDreamer}} \\  
 \rowcolor{gray!10}PhysGaussian~\cite{xie2024physgaussian} & 4.50 & 7.56 & - \\
 PhysDreamer~\cite{zhang2024physdreamer} & 4.54 & 7.71 & -  \\ 
 \rowcolor{gray!10}Physics3D~\cite{liu2024physics3d} & 4.62 & 7.83 & 1.5h \\
 Feature Splatting~\cite{qiu2024feature} & 4.53 & 7.49 & - \\
 \rowcolor{gray!10}DreamGaussian4D~\cite{ren2023dreamgaussian4d} & 4.57 & 7.28 & 0.1h \\
 \rowcolor[HTML]{D7F6FF} \textbf{\our} & \textbf{4.66} & \textbf{7.89} & \textbf{~2min} \\
\hline \hline

\multicolumn{4}{c}{\textbf{synthesized dataset}} \\  
 \rowcolor{gray!10}PhysGaussian~\cite{xie2024physgaussian} & 4.94 & 7.35 & - \\
 DreamPhysics~\cite{huang2024dreamphysics} & 5.05 & 7.92 & 1.5h  \\ 
 \rowcolor{gray!10}Physics3D~\cite{liu2024physics3d} & 5.10 & 8.01 & 1.5h \\ 
 Feature Splatting~\cite{qiu2024feature} & 4.99 & 7.27 & - \\
 \rowcolor{gray!10}DreamGaussian4D~\cite{ren2023dreamgaussian4d} & 4.98 & 6.81 & 0.1h \\
 \rowcolor[HTML]{D7F6FF} \textbf{\our} & \textbf{5.10} & \textbf{8.20} & \textbf{~2min} \\  
\hline \hline

\multicolumn{4}{c}{\textbf{Open-world dataset}} \\  
 \rowcolor{gray!10}Feature Splatting~\cite{qiu2024feature} & 4.79 & 7.10 & - \\
\rowcolor[HTML]{D7F6FF} \textbf{\our} & \textbf{4.96} & \textbf{7.99} & \textbf{~2min} \\  
 \end{tabular}
\vspace{-9pt}
 \caption{\textbf{Quantitative comparisons} on PhysDreamer~\cite{zhang2024physdreamer}, synthesized dataset~\cite{liu2024physics3d}, and Open-world dataset~\cite{kerr2023lerf,haque2023instruct}. 
 RS (Realism Score) represents user study ratings, while AS (Aesthetic Score) is estimated using the LAION aesthetic predictor. Time measures the inference speed for physics-based 4D generation on an single RTX 4090 GPU.}
 \label{tab:exp1}
\end{table}


\subsection{Comparison with SOTA Methods}
\label{sec:exp}
We chose the performance from real-world static scenes from PhysGaussian~\cite{xie2024physgaussian} for fair comparison. 
Tab.~\ref{tab:exp1} presents the user study results (RS) and aesthetic score (AS) predicted by LAION aesthetic predictor following~\cite{huang2024vbench}. Since PhysDreamer~\cite{zhang2024physdreamer} has not released its training code, we only compare the four evaluation scene and are unable to report its inference time. PhysDreamer~\cite{zhang2024physdreamer} scores lower than DreamGaussian4D~\cite{ren2023dreamgaussian4d} in RS and PhysGaussian~\cite{xie2024physgaussian} in AS, which indicates that pre-generated videos may not be a proper ground truth for supervision. Feature Splatting~\cite{qiu2024feature} categorizes objects into types like ``elastic'', ``snow'', ``sand'', and ``water'', directly assigning predefined physical properties, which limits its ability to handle the diverse range of objects.  \textit{Our \our achieves better performance in both metrics}, which demonstrates that \our generates videos that are both realistic and physically plausible, with a high degree of naturalness.

Following~\cite{zhang2024physdreamer}, we also compare the results with real captured videos in Fig.~\ref{fig:results_dreamer}. We utilize space-time slices to present our comparisons, which depict time along the vertical axis and spatial slices of the object along the horizontal axis, as indicated by the red lines in the “object” column. Through these visualizations, we aim to elucidate the magnitude and frequencies of the oscillating motions under scrutiny. \our generates smooth and realistic motion patterns, accurately capturing the natural flow and details of real-world movements. \textit{Please see our project website videos for more video visualization}.

We also evaluate our \our using the synthesized dataset~\cite{liu2024physics3d}. We report the quantitative results against recent methods~\cite{xie2024physgaussian,huang2024dreamphysics,liu2024physics3d,ren2023dreamgaussian4d} in Tab.~\ref{tab:exp1}. Our method still generates the most consistent and natural motions. The visual results are shown in Fig.~\ref{fig:results_3d}.




\begin{table}[t]
 \centering
 \setlength{\tabcolsep}{8pt}
 \small
  \begin{tabular}{cccccc}
 \toprule
 GPT & BLIP & CLIP & w/o MPDP & PGAS & AS
 \\ 
 \hline
  \rowcolor{gray!10}\checkmark &  &  & \checkmark & \checkmark & 4.47 \\
  \checkmark & \checkmark &  & \checkmark & \checkmark & 4.59  \\
  \rowcolor{gray!10}\checkmark & \checkmark & \checkmark &  & \checkmark & 4.64 \\
  \checkmark & \checkmark & \checkmark & \checkmark &  & 4.62 \\
  \rowcolor{gray!10}\checkmark & \checkmark & \checkmark & \checkmark & \checkmark & \textbf{4.66} \\
 \end{tabular}
 \vspace{-7pt}
 \caption{\textbf{Ablation Study} on PhysDreamer~\cite{zhang2024physdreamer} dataset. AS denotes the average aesthetic quality score predicted using the LAION aesthetic predictor.}
 \label{tab:ablation}
\end{table}

\subsection{Ablation study}
\label{sec:ablation}

In this section, we conduct ablation experiments using PhysDreamer~\cite{zhang2024physdreamer} dataset to evaluate the effectiveness of our proposed modules. 

\begin{figure}[t]
  \centering
    \includegraphics[width=0.95\linewidth]{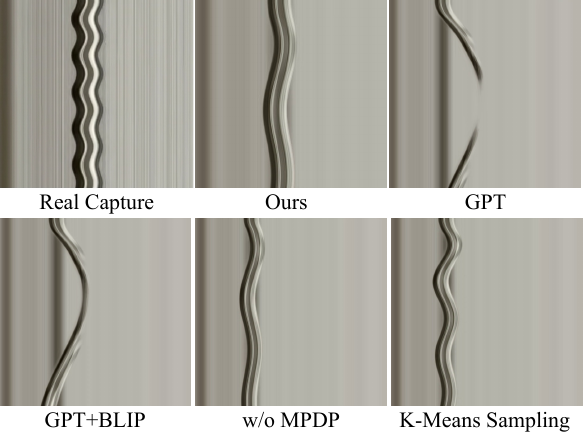}\\
      \vspace{-9pt}
  \caption{\textbf{Ablation study}. Visualization of space-time slices for ablation study on PhysDreamer~\cite{zhang2024physdreamer}. Our method can generate closer content compared with the real capture.}
  \label{fig:ablation}
\end{figure}

\medskip
\noindent
\textbf{Model for physical property perception.} In Fig.~\ref{fig:ablation} we compare two methods with our proposed MLLM-P3: 
1) GPT: Predicting physical properties using only the image;
2) GPT+BLIP: Predicting properties with both the image and a text description from BLIP;
3) GPT+BLIP+CLIP (MLLM-P3): Generating a dictionary of K candidate materials with GPT, selecting the best match via CLIP, and then predicting properties using the image, description, and chosen material.
As shown in Tab.~\ref{tab:ablation}, MLLM-P3 performs best because there is inherent uncertainty in predicting materials based on just visual appearance or text description, as shown in Tab.~\ref{tab:ablation}, .

\medskip
\noindent
\textbf{Material property distribution prediction.} Material property distribution prediction is designed for complex physical properties distribution. As shown in Fig.~\ref{fig:ablation} and Tab.~\ref{tab:ablation}, it is required to achieve optimal performance.

\medskip
\noindent
\textbf{Sampling strategy selection.} 
Compared to PhysDreamer's K-Means sampling~\cite{zhang2024physdreamer}, our PGAS strategy generates more temporally coherent results, as shown in Fig.~\ref{fig:ablation} with significantly improved visual metrics in Table~\ref{tab:ablation}. This demonstrates PGAS's enhanced capability in reconstructing photorealistic 4D dynamics.

\section{Extension to Open-world Scene}
\label{sec:extension}
To evaluate the physical simulation accuracy in open-world 3D scenes, we select some multi-object scenes from~\cite{kerr2023lerf,haque2023instruct}. Our method simulate multiple objects simultaneously in open-world scenarios, unlike current methods ({\em e.g.}, PhysDreamer~\cite{zhang2024physdreamer} and Physics3D~\cite{liu2024physics3d}) that handle only single objects. Therefore, we omit direct comparisons with these methods in open-world settings. Qualitative results are shown in Fig.~\ref{fig:rebuttal} and on our project page, with quantitative results in Tab.~\ref{tab:exp1}. Additionally, we integrate LGM~\cite{tang2025lgm} to generate 3D assets from single images and perform physical simulations as demonstrated in Fig.~\ref{fig:rebuttal}.

\section{Conclusion}
\label{sec:con}
We introduce \our, a framework for generating physics-based dynamics and photorealistic renderings.
We begin with precise scene reconstruction and object-level 3D open-vocabulary segmentation, followed by multi-view image in-painting. Then, we propose MLLM-based Physical Property Perception (MLLM-P3) to predict mean physical properties of objects. Using these mean values and object geometry, the Material Property Distribution Prediction model (MPDP) then estimates the complete distribution, reframing the task as probability distribution estimation to reduce computational costs. Finally, we simulate objects with particles sampled via the Physical-Geometric Adaptive Sampling (PGAS) strategy. Extensive experiments and user studies show that \our produces more realistic motion than state-of-the-art methods within much faster inference time.
We believe that \our represents a meaningful advance toward more engaging and immersive virtual environments, unlocking diverse applications from realistic simulations to interactive virtual experiences.

\medskip
\noindent
\textbf{Limitation and future work.} 
\our struggles to segment entire objects with occluded objects, leading to unnatural simulations. Future work will explore to reconstruct occluded parts, further enhancing realism and expanding applications in interactive virtual experiences.



\medskip
\noindent
\textbf{Acknowledgement}
This work was supported by Central Guidance for Local Science and Technology Development Fund (ZYYD2025QY19), and the Bingtuan Science and Technology Program (2022DB005).

{
    \small
    \bibliographystyle{ieeenat_fullname}
    \bibliography{main}
}


\end{document}